\documentclass[runningheads]{llncs}
\usepackage{graphicx}
\usepackage{amsmath}
\usepackage{amssymb}
\usepackage{url}
\usepackage{float}
\usepackage{caption}

\usepackage{subcaption}
\usepackage{booktabs}

\usepackage{array}

\DeclareMathOperator*{\argmax}{arg\,max}

\begin{document}
\title{Anomaly Detection for imbalanced datasets with Deep Generative Models}
%
%
\author{Nazly Rocio Santos Buitrago \inst{1} \and
Loek Tonnaer \inst{1} \and
Vlado Menkovski \inst{1} \and Dimitrios Mavroeidis \inst{2}}
\authorrunning{N. Santos et al.}
%
\institute{Eindhoven University of Technology, Eindhoven, The Netherlands \and
Royal Philips B. V., Eindhoven, The Netherlands}
\maketitle              
\begin{abstract}
Many important data analysis applications present with se-verely imbalanced datasets with respect to the target variable. A typical example is medical image analysis, where positive samples are scarce, while performance is commonly estimated against the correct detection of these positive examples. We approach this challenge by formulating the problem as anomaly detection with generative models. We train a generative model without supervision on the `negative' (common) datapoints and use this model to estimate the likelihood of unseen data. A successful model allows us to detect the `positive' case as low likelihood datapoints.

In this position paper, we present the use of state-of-the-art deep generative models (GAN and VAE) for the estimation of a likelihood of the data. Our results show that on the one hand both GANs and VAEs are able to separate the `positive' and `negative' samples in the MNIST case. On the other hand, for the NLST case, neither GANs nor VAEs were able to capture the complexity of the data and discriminate anomalies at the level that this task requires. These results show that even though there are a number of successes presented in the literature for using generative models in similar applications, there remain further challenges for broad successful implementation.

\end{abstract}
\section{Introduction}
A long-standing challenge for Machine Learning is to deal with small datasets and an insufficient amount of labeled data\cite{Ultrasound_smallData}. This is particularly true when there is a significant imbalance in the data with respect to the class (or target variable). 
We address this challenge by formulating it as an anomaly detection task. Specifically, we train a generative model in an unsupervised fashion with the samples from only one class. We treat the other class as an anomaly, such that our model is expected to produce low likelihood of samples from the other class. 

In other words, we consider a Probability Density Estimation process in which the goal is to discover the probability distribution of the \textit{normal} data $p_{data}$, by defining a parametric distribution $p_{model}$ and finding the optimal parameters to approximate $p_{data}$. Computing these optimal parameters $\theta$, means getting the values that maximize the likelihood of the observed data.

Given a set of training datapoints $X = \{x_1,x_2,...,x_n\}$, we train a generative model to learn the probability distribution $p(x)$. The model inference is based on Maximum Likelihood Estimation (MLE) for the parameters $\theta$. Having the likelihood function $p_{model}(X|\theta)$, the MLE is defined by:

\begin{equation}{\label{equation:MLE}}
    \theta_{MLE} =  \argmax_\theta p_{model}(X|\theta)\\
    = \argmax_\theta \prod_i p_{model}(x_i|\theta)
\end{equation}

When computing the MLE, we find the parameters that maximize the likelihood of the data given our model $p_{model}$. 

More specifically, we define the optimization as a minimization of a negative log likelihood given by\cite{MLEToLoss}: 

\begin{equation}{\label{equation:NegLogLik}}
E(w) = -\sum_i \log p[x_i|f(x_i;w)],
\end{equation}

where the model $f(.;w)$ is a type of neural network with parameters $w$ defined by the specific generative model. By having this form, the task becomes an optimization process that can be solved using Stochastic Gradient Descent (SGD). 

Furthermore, we need to develop a boundary to distinguish anomalies by developing a threshold $\epsilon$, with respect to the learned likelihood. It is also the case that this cut-off is not obvious to identify and relies entirely on experts' opinions\cite{LOCI_density_outlier}. 

Deep Generative Models are the current unsupervised methods with strong capacity for feature representation, data generation and learning of the data distribution. Their structure, using neural networks, allows them to construct powerful functions from the training and generate new \textit{alike} samples, particularly for high dimensional data, for which density estimation is a long standing problem.  

Our particular goal is to apply this approach to difficult applications such as lung cancer screening. Lung cancer alone was responsible for 1.69 million deaths in 2015\footnote{\url{http://www.who.int/news-room/fact-sheets/detail/cancer}}. Early cancer detection and diagnosis of abnormal anatomies, by means of Computer Tomography (CT), has been a recurrent research topic specially in the Computer Vision domain\cite{CADe_medicalImages}\cite{CT_deepLearning}. 

\section{Related Work}\label{section:relatedwork}
Two main frameworks gained popularity and acceptance in the deep learning community: Generative Adversarial Networks\cite{Goodfellow2014} (GAN) and Variational AutoEncoders\cite{VAEs_welling} (VAE). Since their appearance in 2013-2014, strong research moved into their interpretation, application and development. Currently there are more 200 variations of GANs in terms of training, architecture, loss function, objective and applications\footnote{\url{https://github.com/hindupuravinash/the-gan-zoo}}. 
GANs are known for being unstable to train, with several hyper-parameters to tune. However, the results are sharp, and could fool the human eye when producing new image samples. VAEs are known for producing blurry results in the new samples. However, their training setup is well defined, and they feature explicit sampling from the learned probability distribution. Due to their proven performance when dealing with high-dimensional datasets in an unsupervised setup, deep generative models are suitable for the design of the anomaly detection framework. 

Other recent and important density estimation methods are the autoregressive model (with normalizing flows), the Neural Autoregressive Distribution Estimator (NADE)\cite{NADE_autoregressiveDistri_estimator_2011}, the real-valued neural autoregressive density-estimator (RNADE)\cite{RNADE_autoregressiveDistri_estimator_2013}, the real-valued non-volume preserving method\cite{RealNVP_estimation_2016}, and the Masked Autoregressive Flow estimation\cite{masked_autoregressiveFlow_2017_artstate}. An improved VAE approach with inverse autoregressive flows\cite{vae_inverseflow} has also demonstrated strong capacity for density estimation. In the current version of the work we have not evaluated the autoregressive flow models.    
    
\section{Applications}\label{section:applications}

\subsection{MNIST dataset, 2D benchmark setup}
We presented an evaluation to the anomaly detection over the benchmark dataset MNIST\footnote{\url{http://yann.lecun.com/exdb/mnist/}}. For our experiments we split the dataset in a binary classification problem, having an imbalanced setup. 
We trained only using the Negative Samples (a subset of 9216 images containing equal samples of {{0-8}}). Then we tested the approach using some Positive samples (images of {{9}}). 

\subsection{Lung cancer detection, nodules from NLST 3D dataset}
\begin{figure}[ht!]
    \centering
\begin{subfigure}{0.4\textwidth}
  \centering
  \includegraphics[scale=0.6]{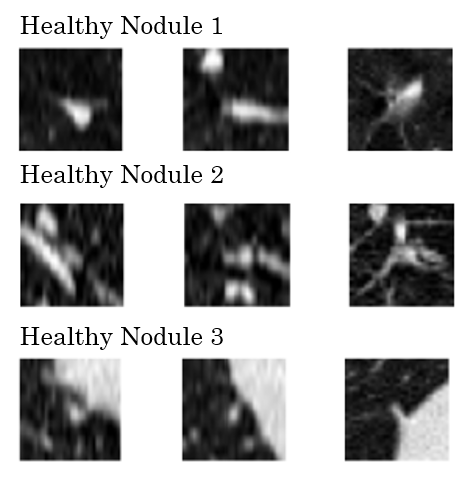}
  \caption{Healthy Samples}
  \label{fig:healthysample}
\end{subfigure}
\begin{subfigure}{0.4\textwidth}
  \centering
  \includegraphics[scale=0.6]{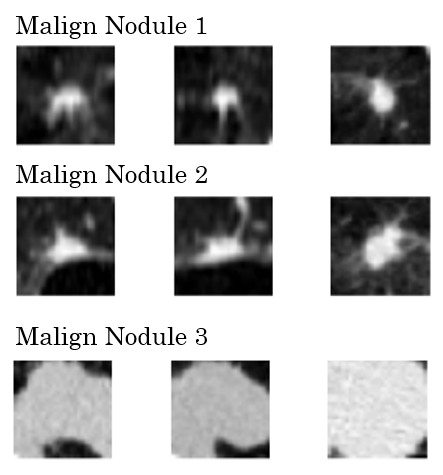}
  \caption{Abnormal Samples}
  \label{fig:cancersample}
\end{subfigure}%
\caption{Examples of samples in the dataset with their axial, coronal and sagital perspective. Figure \ref{fig:healthysample} shows 3 different healthy nodules. Figure \ref{fig:cancersample} shows 3 different nodules identified as abnormal (positive for cancer). }
\label{fig:Samples}
\end{figure}

Lung cancer detection usually requires annotated images (cancer, non-cancer) at a nodule (tumor) level, with its additional information such as malignancy, diameter, spiculation or lobulation, and a preferably amount of samples of each class.  Recent efforts\footnote{\url{https://github.com/dhammack/DSB2017/blob/master/dsb_2017_daniel_hammack.pdf}} leveraged from the use of the publicly available datasets with considerable nodule annotations, achieving good performance. However, this supervised approach does not seem to be easily scalable due to the lack of new, equally rich data. 
In this particular application, the benign nodules of the lung do not share specific characteristics. They are diverse in size, texture, shape, and location. As a consequence, the differentiation between malign nodules is not evident for human perception. Due to the high complexity of the data, we are not sure how far the abnormal samples are from the normal samples. We would like to test our anomaly detection framework in this scope and evaluate whether the generative models are able to understand class related particularities such as shapes, edges or spatial position, plus additional hidden features.
The raw dataset is provided by the NLST (National Lung Screening Trial), consisting of high resolution chest tomographies. The input for our models is the result of a nodule detector. We are dealing with 3D cubes of 32x32x32 mm\textsuperscript{3} with a voxel size of 1mm\textsuperscript{3}. In our research we designed and implemented 3D models for handling the data, and investigated whether this approach helped for the robust estimation of the probability density. Figure \ref{fig:Samples} shows some nodule examples, the variation between the data and the difficulty for humans to discriminate healthy from abnormal samples. 

\begin{figure}[ht!]
\centering
\includegraphics[scale=0.5]{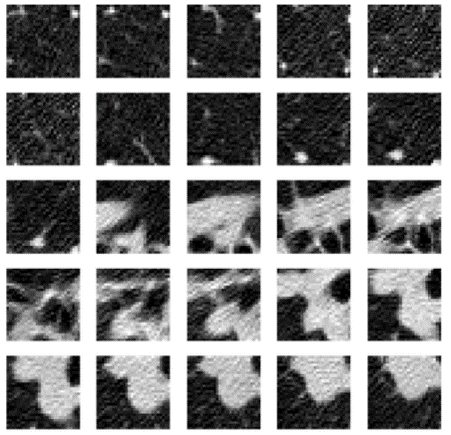}\\
\caption{Displaying 25 slices of 28x28 pixels, as a representation of the cube of 28x28x28 pixels used for training the models.}
\label{fig:1.healthysample3d}
\end{figure}

For our experiments the input of the models is a 3D cube of 28x28x28 pixels, the result of a data augmentation process that produces sub patches from the original shape of 32x32x32. For convenience in display, figure \ref{fig:1.healthysample3d} shows the 3D image as a set of 25 slices of 28x28 pixels. Table \ref{table:NLSTdataset} shows the details of the how we organized the data.  
\begin{table}[ht!]
\centering
\begin{tabular}{@{}llll@{}}
    \toprule
               & \begin{tabular}[c]{@{}l@{}}Label 0\\ Healthy\end{tabular} & \begin{tabular}[c]{@{}l@{}}Label 1\\ Cancer\end{tabular} & Total \\ \midrule
    Training   & 1722                                                      & 460                                                      & 2182  \\
    Validation & 431                                                       & 115                                                      & 546   \\
    Testing    & 539                                                       & 143                                                      & 682   \\ \bottomrule
\end{tabular}  
\caption{Lung nodule dataset after data augmentation}
\label{table:NLSTdataset}
\end{table}

\section{Anomaly Detection Framework}
\subsection{Anomaly Detection with GANs}
The reference paper for Anomaly Detection\cite{AnoGAN}, based on work from\cite{Raymond_2GAN}, proposed a framework composed of three steps: (1) learn a manifold $\mathcal{X}$ of a corpus of \textit{normal} images, (2) map images back to the latent space, and (3) detect abnormal samples using a visual and perceptual component.

The model used to learn the manifold for step (1) is a GAN\cite{Goodfellow2014}, consisting of a generator $G$ that generates images given latent space samples $z$, and a discriminator $D$ that is trained to distinguish generated images from real data. Both $G$ and $D$ are neural networks.

Mapping an image $x$ back to the latent space in step (2) entails finding some $z_\gamma$ in latent space such that $G(z_\gamma)$ is as similar as possible to $x$.

The visual component of step (3) is the residual loss, which compares similarity of images at pixel level through the generator $G$. The residual loss is defined by

\begin{equation}\label{equation:residualLoss}
 \mathcal{L}_R(z_\gamma) = \sum|x - G(z_\gamma)|,
\end{equation}

where $x$ is the query image and $G(z_\gamma)$ is the most similar generated image. If the generator is able to generate a perfect looking image with respect to the query, the residual loss is $\mathcal{L}_R(z_\gamma)=0$.
The perceptual component is defined as a discriminator loss, based on the discriminator $D$:
\begin{equation}\label{equation:perceptualLoss}
 \mathcal{L}_D(z_\gamma) = \sum|f(x) - f(G(z_\gamma))|,
\end{equation}

where $f$ is a hidden layer from the discriminator. The features learned from the query image $f(x)$ are compared to the ones of the most similar generated image $f(G(z_\gamma))$. 

The method for detecting an abnormal sample consists of using the overall loss composed by a weighted sum of the residual and the discriminator loss. A parameter $\lambda$ sets the relative importance of each loss component:

\begin{equation}\label{equation:AnoTotalLoss}
 \mathcal{L}(z_\gamma) = (1-\lambda)\mathcal{L}_R(z_\gamma) + \lambda\mathcal{L}_D(z_\gamma).
\end{equation}

An iterative procedure\cite{Raymond_2GAN} is used to find $z_\gamma$; starting with a random point $z_1$ in latent space that generates an image $G(z_1)$, and then using equation~\ref{equation:AnoTotalLoss} to find more suitable $z_2, z_3, \ldots, z_\gamma$ through stochastic gradient descent (SGD) with momentum. After $\gamma$ steps, if the query image $x$ belongs to the learned distribution of the model, we would expect $G(z_\gamma) \approx x$.  
 
After training, we obtain the closest image to the query $x$, generated by $G(z_\gamma)$ and the loss value $\mathcal{L}(z_\gamma)$. As suggested in the paper, we use equation \ref{equation:AnoTotalLoss} to set a threshold $\epsilon$ on $\mathcal{L}(z_\gamma)$ for Anomaly Detection. The reasoning is that if the query image $x$ is close to the learned representation, it is consider \textit{normal} and will have a low loss. If $x$ is \textit{abnormal}, it will have a higher loss, above the defined threshold.

\subsection{Anomaly Detection with VAEs}
A Variational Autoencoder\cite{VAEs_welling} (VAE) is a latent variable model that uses neural networks to express the parameters $\phi$ of an approximate posterior distribution $q_\phi(z|x)$ over the latent variables $z$ (the encoder), as well as for the parameters $\theta$ of a generative model $p_\theta(x|z)$ (the decoder), given some prior distribution $p(z)$ for the latent variables. It is trained on maximizing the Evidence Lower Bound (ELBO), a lower bound to the log likelihood $\log p(x)$ of the data. The ELBO can be formulated as:
\begin{equation}\label{eqn:elbo}
ELBO(\phi, \theta; x) = \mathbb{E}_{q_\phi(z|x)}[\log p_\theta(x|z)] - KL(q_\theta(z|x)||p(z)),
\end{equation}
where $KL(\cdot||\cdot)$ is the KL divergence between two probability distributions. The first term in equation~\ref{eqn:elbo} can be interpreted as a reconstruction error on pixel level, whereas the second term acts as a regularizer. We also use this ELBO as an approximation to the likelihood, for use in our anomaly detection framework.

In our experiments, we use multivariate Gaussians with diagonal covariance:
\begin{align}
q(\pmb{z}|\pmb{x}) &= \mathcal{N}(\pmb{z} | \pmb{\mu}_{enc}(\pmb{z}), \pmb{\sigma}_{enc}(\pmb{z})), \\
p(\pmb{x}|\pmb{z}) &= \mathcal{N}(\pmb{x} | \pmb{\mu}_{dec}(\pmb{x}), \sigma_{dec} \cdot \pmb{I}), \\
p(\pmb{z}) &= \mathcal{N}(\pmb{z} | \pmb{0}, \pmb{I}).
\end{align}

In this case, the KL divergence term from equation~\ref{eqn:elbo} can be computed analytically, whereas the expectation in the first term can be approximated efficiently by means of Monte Carlo sampling. During training we use a single sample, but for evaluation of the ELBO in our anomaly detection framework we sample 100 instances from the approximate posterior, in order to find a reliable estimate of the likelihood of a data point. We used a fixed value $\sigma_{dec} = \frac{1}{\sqrt{2}}$, whereas $\pmb{\mu}_{enc}(\pmb{z}), \pmb{\sigma}_{enc}(\pmb{z}),$ and $\pmb{\mu}_{dec}(\pmb{x})$ are all expressed by neural networks.

Similar approaches\cite{AnoVAEGAN2018_MRI2D} use only the reconstruction error (the first term in equation~\ref{eqn:elbo}) for anomaly detection, from the perspective of image segmentation. However, by using just that part of the VAE loss function, we are not really estimating the true likelihood $p(x)$. The goal of our Anomaly Detection framework is to estimate how likely it is that an image query belongs to the learned distribution.

\subsection{Evaluation}
After training the generative models, we expect that the model learned specific features from the data and the resultant loss values can be seen as a likelihood value for each data point, measuring how likely it is for that sample to belong to the distribution of normal data. We then expect that the likelihood of normal samples is far greater than for anomalous data. 
To evaluate this assumption, we took an equal number of \textit{normal} and \textit{anomaly} samples and computed the likelihood value for all the datapoints, using a trained GAN and VAE.


\section{Results}
We present results for the Anomaly Detection framework with GAN architectures (GAN-AD) and VAE architectures (VAE-AD). For both phases the tests were performed for the application cases described in section~\ref{section:applications}, over the 2D MNIST and 3D NLST datasets. As a general structure, we present:  
\begin{itemize}
    \item High level defined architecture for the GAN-AD and VAN-AD over the NLST dataset,
    \item Qualitative performance of the models in terms of the generation/reconstruc-tion of samples,
    \item Visual evaluation of the Anomaly Detector output using plot of density distributions,
    \item The AUROC score obtained after thresholding the Anomaly Detector output to separate the anomalies based on their likelihood measure.
\end{itemize}

\subsection{MNIST 2D setup}
For both models we trained with a subset of 9216 images containing normal samples, images of digits 0 to 8. Then we evaluated the approach using an equal number of samples from both classes: 450 positive samples (images of number {{9}}) and 450 additional normal samples.

\subsubsection{GAN-AD}
For the MNIST dataset dimensions, we used the proposed DCGAN\cite{Radford_DCGAN}, with similar configuration of the convolutional layers, and the same recommendations for training. These type of implementations are widely explored by the community, so there was no need to tune the hyper-parameters with rigor. We trained for 50 epochs and we computed the anomaly scores using 100 backpropagation steps for finding the optimal $z$ mapping back to latent space. We chose $\lambda=0.5$ in equation \ref{equation:AnoTotalLoss}, after empirical experimentation. 

\begin{figure}[ht!]
    \centering
    \includegraphics[scale=0.4]{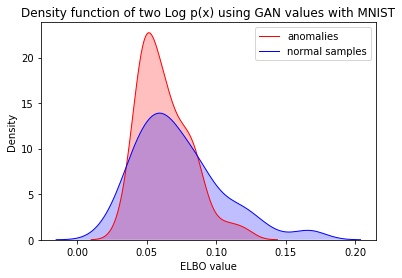}
    \caption{Density distribution for the anomaly scores obtained with GAN-AD with MNIST dataset. We can perceive that there is no clear separation or threshold between normal and abnormal samples.}
    \label{fig:gan_mnist_density}
\end{figure}

After the training, we performed the Anomaly Detection framework, obtaining the anomaly scores (equation~\ref{equation:AnoTotalLoss}). Figure~\ref{fig:gan_mnist_density} shows the distribution of the scores corresponding to each class. We found high fragility in the results when we: increased the trained epochs, increased the number of backpropagation steps for finding the optimal $z$ in latent space, and, changed the number of samples used both for training and for evaluation. 

\begin{figure}[ht!]
    \centering
    \includegraphics[scale=0.4]{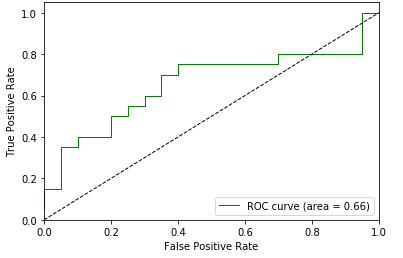}
    \caption{ROC curve with AUROC score for GAN-AD with MNIST.}
    \label{fig:gan_mnist_auroc}
\end{figure}

For quantitative evaluation, we plot a ROC curve based on thresholding the anomaly score on different values. The result is shown in figure \ref{fig:gan_mnist_auroc}. The Area Under Curve (AUROC) value of 0.66 shows that the classifier is somewhat able to separate normal from abnormal samples, although not in a powerful way. As explained before, this value was fragile for the number of samples used in the evaluation and the previous training.

\subsubsection{VAE-AD}

For the 2D context of MNIST, we use a simple VAE architecture with 2D convolutions and 2D upsamplings. For this dataset there was no need for a deep level of convolutions or number of units. We trained the VAE for 30 epochs, using the same 9216 images labeled as \textit{normal} (digits from 0 to 8). The training lasted approximately 1 min. The model is able to visually reconstruct a normal sample with high quality, and tries to approximate an abnormal sample with the information it got during training.

\begin{figure}[ht!]
    \centering
    \includegraphics[scale=0.4]{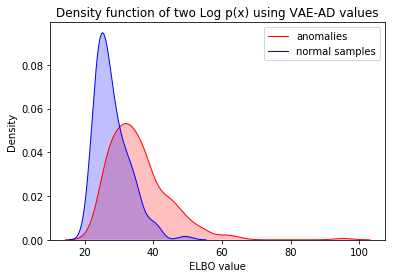}
    \caption{Density distribution of VAE-AD with MNIST dataset. We can see how the values create a differentiation in the densities. Values greater than 50 are highly probable to be anomalies.}
    \label{fig:5.3DVAE_likelihoods}
\end{figure}

The metrics of our model were used for the computation of the likelihood lower bound. For the VAE-AD, we took the trained VAE and passed new samples through it. For this step we used 450 normal samples and 450 abnormal samples. 
The result is a density graph composed by the ELBO values. Figure~\ref{fig:5.3DVAE_likelihoods} shows the distribution of the results for both types of samples. 

\begin{figure}[ht!]
    \centering
    \includegraphics[scale=0.4]{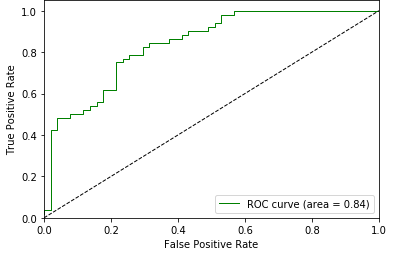}
    \caption{ROC curve with AUROC score for VAE-AD for MNIST. The result implies that the framework has the potential to discriminate normal from abnormal samples.}
    \label{fig:5.3DVAE_AUROC}
\end{figure}

Figure \ref{fig:5.3DVAE_AUROC} shows the ROC curve and AUROC score results based on thresholding the ELBO scores on different values. The AUROC score of 0.84 shows high potential for differentiation between normal and anomalous samples.
\subsection{NLST 3D nodules dataset}
\subsubsection{GAN-AD}
\label{section:gan-adResult}
\begin{figure}[ht!]
    \centering
    \includegraphics[scale=0.6]{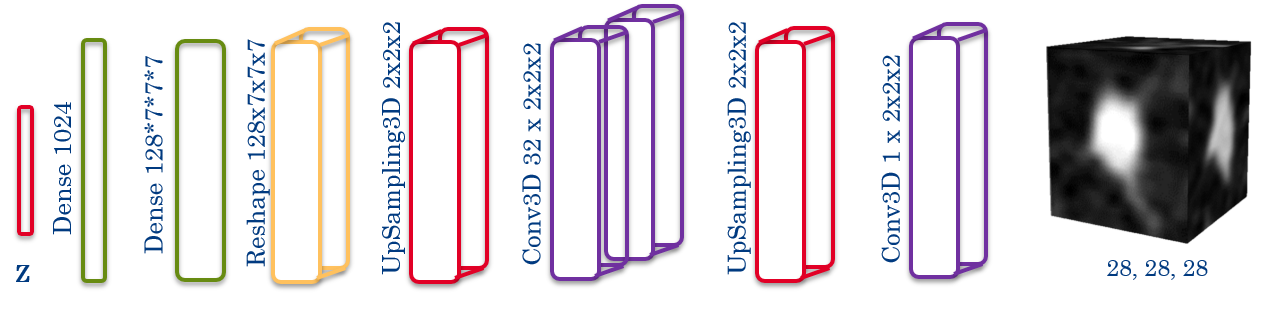}
    \caption{Trained 3D WGAN-GP architecture for Generator}
    \label{fig:5.3DGAN_arc_2}
\end{figure}

After exhaustive parameter tuning and attempts in training, figure \ref{fig:5.3DGAN_arc_2} shows the 3D WGAN-GP\cite{WGAN-GP} architecture that was able to learn from the nodule data and produced some visually understandable results.

\begin{figure}[ht!]
\centering
\includegraphics[scale=0.6]{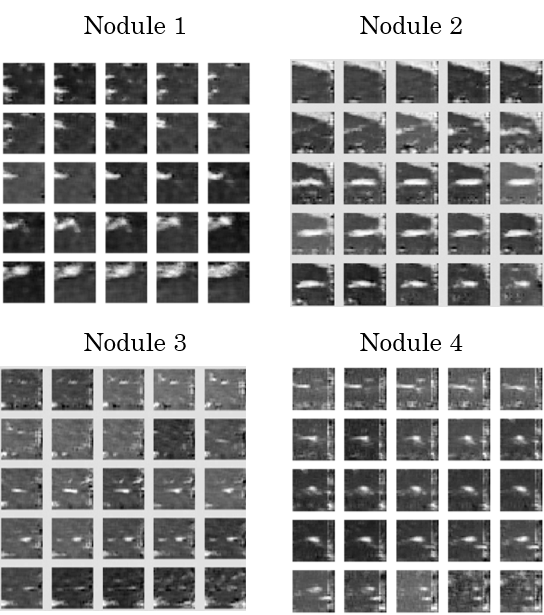}
\caption{Four nodules generated by the 3DGAN}
\label{5.GAN_3D_samples}
\end{figure}

Training was configured with a seed $z$ of size 100 following a uniform distribution. We trained for 100 epochs, since the loss function for the critic showed optimization around epoch 50 and stops learning from epoch 60. The samples, however, keep improving visually until 100 epochs. Figure \ref{5.GAN_3D_samples} shows examples of new data produced by the GAN.

While comparing different generated images from the variations in training of WGAN-GP, we notice a partial mode collapse\cite{GANTutorial} in the samples. We can see that the images look similar and they are not able to create complex shapes as seen in the training data. This was the case for less sharp images generated from simpler architectures or with change on the random seed $z$. Even when the generator is able to construct simple shapes, they are very similar to each other. 

\begin{figure}[ht!]
\centering
\includegraphics[scale=0.4]{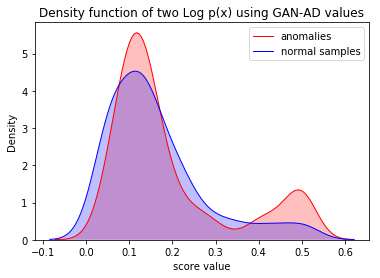}
\caption{Density distribution of values using GAN-AD with NLST}
\label{fig:5.3DGAN_density}
\end{figure}

With the final trained architecture as shown in figure~\ref{fig:5.3DGAN_arc_2}, we compute the metrics proposed in our methodology.

We used the test split, 120 normal samples and 120 abnormal samples, for calculation of the loss score. We ran 100 backpropagation steps for mapping images into the latent space, and we chose $\lambda=0.5$ in equation~\ref{equation:AnoTotalLoss}, after empirical experimentation. The experiment setup showed that backpropagating in the latent space was resource consuming, taking almost 30 seconds per image for 100 steps. Also, giving more weight $\lambda$ to one loss did not improve the resulting optimization. Figure \ref{fig:5.3DGAN_density} shows the distribution of the results. Visually, it is clear that the model is not able to differentiate the distribution of normal samples from abnormal, as they overlap. 
\begin{figure}[ht!]
    \centering
    \includegraphics[scale=0.4]{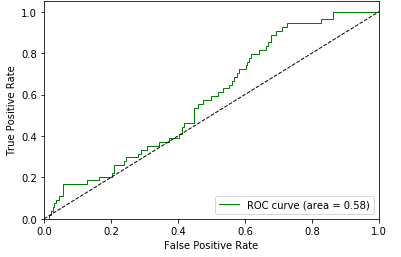}
    \caption{ROC curve with AUROC score for GAN-AD with NLST. The result implies the classifier was not able to discriminate any feature from normal to abnormal samples.}
    \label{fig:5.3DGAN_AUROC}
\end{figure}

\begin{figure}[ht!]
    \centering
    \includegraphics[scale=0.4]{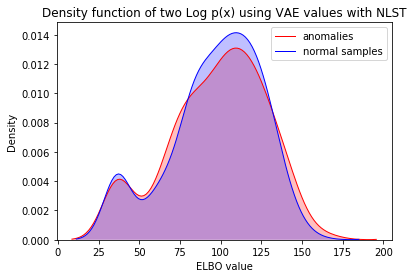}
    \caption{Density distributions of VAE-AD outputs for NLST dataset}
    \label{fig:vaenlstDensities}
\end{figure}

Figure \ref{fig:5.3DGAN_AUROC} shows the ROC curve and AUROC score results. A value of 0.58 implies that the input features were not relevant enough, and the classifier was able to perform just better than random guessing.

\begin{figure}[ht!]
    \centering
    \includegraphics[scale=0.4]{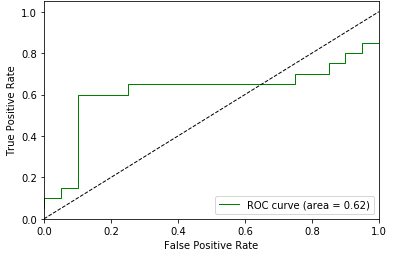}
    \caption{ROC curve with AUROC score for VAE-AD with NLST. The result implies the classifier performs better than random choice, but still, it does not have the capacity for discriminating normal from abnormal samples.}
    \label{fig:vaenlstAUROC}
\end{figure}


\subsubsection{VAE-AD}
\begin{figure}[ht!]
    \centering
    \includegraphics[scale=0.5]{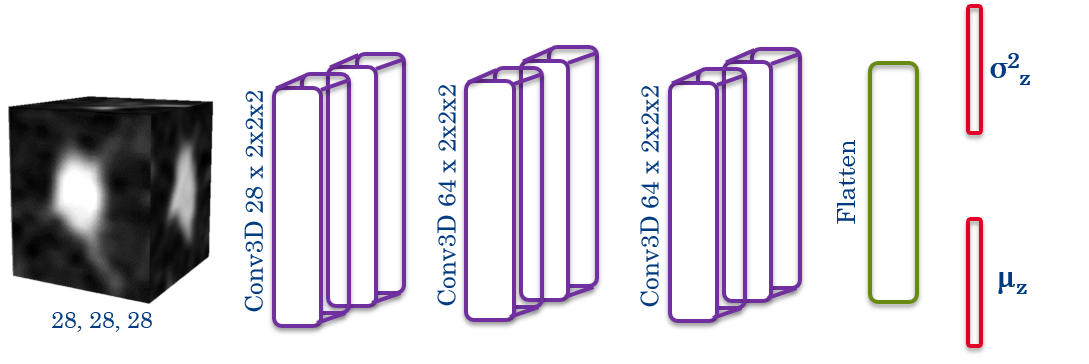}
    \caption{Trained 3D VAE architecture for Encoder}
    \label{5.3DVAE_arch}
\end{figure}

Based on the performance of the 3D WGAN-GP architecture, we trained a 3D VAE using a similar setup of 3D convolutional layers and Upsampling3D. Figure \ref{5.3DVAE_arch} shows the architecture used for the encoder.

Using the same 1722 normal nodules as for GANs, we trained the model for 100 epochs. As for the VAE-AD, we used the resultant metrics for computation of the likelihood lower bound. We used the trained 3D VAE and passed new samples through it. We used 115 normal samples and 115 anomaly samples. The distribution of the values is shown in figure~\ref{fig:vaenlstDensities}. Visually it is clear that the distributions overlap, not making an ideal separation between normal and anomalous samples.  

Figure~\ref{fig:vaenlstAUROC} shows the resulting ROC curve for thresholding on different values. We can see that even if we perform better than random guessing, the given representation was not enough to make a clear distinction between normal and abnormal samples. Empirically, we noticed that increasing the number of samples could improve this score. We used samples from the validation split to perform more experiments, but the AUROC score was not greater than 0.62. In presence of additional data, more experimentation could give better performance.   

\section{Conclusion}
This work defined a comparative Anomaly Detection framework for two state-of-the-art deep generative models. We used a metric based on likelihood estimation, and created an evaluation protocol for the identification of anomalies. The concept of likelihood estimation is closer related to the VAE framework. GAN computes a loss score that has no direct link to probability theory, but that can be interpreted as an anomaly score.

For the first use case with MNIST, the GAN approach is fragile and it is dependent on hyperparameter tuning and the number of training samples. When evaluated with ROC curves, it did not show the expected performance as for the reference paper\cite{AnoGAN}. Results showed an AUROC score of 0.66 when training with fewer than 10.000 samples of the normal class. Since our scope was imbalance and scarcity of samples, this was a realistic scenario for evaluating the model. Regarding the VAE approach, it is easy to train, not time consuming and the scores are obtained in a straightforward manner. The resulting ROC curve shows potential for the separation of abnormal samples, with a value of 0.84.

The use case of lung cancer detection at a 3D image nodule level showed that neither of the generative models are able to capture the feature complexity of the data. The GAN approach evaluation showed a performance just better than random with an AUROC score of 0.58. With a VAE we obtained an AUROC score of 0.62, which we consider not significantly relevant due to the importance of the abnormal samples.

Previous work\cite{Rodrigo2018} showed that GAN-AD did not perform well in an NLST 2D setup. We performed experiments over 3D architectures, expecting a richer model. However, we saw that deep generative models are still not robust enough in cases such as CT data of lung cancer at a nodule level.

\section{Discussion}
The current results showed that deep generative models are a suitable approach for anomaly detection and developing models in highly imbalanced settings. However, their applicability depends on the complexity of the dataset. Particularly, for cancer detection at a nodule level we have yet to develop models that precisely model the distribution of the images to a level that the malignant tumors can be distinguished from the benign lesions.  Recent developments of Autoregressive Models with normalizing flows for density estimation presented in the background section (See section~\ref{section:relatedwork}) offer significant advances to current generative models and are hence a strong candidate for a solution in this domain.

\section*{Acknowledgements}
The authors thank the National Cancer Institute for
access to NCI’s data collected by the National Lung Screening Trial (NLST). The statements contained
herein are solely those of the authors and do not represent or imply concurrence or endorsement by NCI.

%
%
%
\bibliographystyle{splncs04}
\bibliography{references}

\begin{thebibliography}{10}
\providecommand{\url}[1]{\texttt{#1}}
\providecommand{\urlprefix}{URL }
\providecommand{\doi}[1]{https://doi.org/#1}

\bibitem{AnoVAEGAN2018_MRI2D}
Baur, C., Wiestler, B., Albarqouni, S., Navab, N.: Deep autoencoding models for
  unsupervised anomaly segmentation in brain {MR} images. CoRR
  \textbf{abs/1804.04488} (2018), \url{http://arxiv.org/abs/1804.04488}

\bibitem{RealNVP_estimation_2016}
Dinh, L., Sohl{-}Dickstein, J., Bengio, S.: Density estimation using real
  {NVP}. CoRR  \textbf{abs/1605.08803} (2016),
  \url{http://arxiv.org/abs/1605.08803}

\bibitem{GANTutorial}
Goodfellow, I.J.: {NIPS} 2016 tutorial: Generative adversarial networks. CoRR
  (2017), \url{http://arxiv.org/abs/1701.00160}

\bibitem{Goodfellow2014}
Goodfellow, I.J., Pouget-Abadie, J., Mirza, M., Bing~Xu, D.W.F., Ozair, S.,
  Courville, A., Bengio, Y.: {Generative Adversarial Nets}  (2014),
  \url{https://arxiv.org/abs/1406.2661}

\bibitem{CT_deepLearning}
Greenspan, H., van Ginneken, B., Summers, R.M.: Guest editorial deep learning
  in medical imaging: Overview and future promise of an exciting new technique.
  IEEE Transactions on Medical Imaging  \textbf{35}(5),  1153--1159 (May 2016).
  \doi{10.1109/TMI.2016.2553401}

\bibitem{WGAN-GP}
Gulrajani, I., Ahmed, F., Arjovsky, M., Dumoulin, V., Courville, A.C.: Improved
  training of wasserstein gans. CoRR  \textbf{abs/1704.00028} (2017),
  \url{http://arxiv.org/abs/1704.00028}

\bibitem{VAEs_welling}
{Kingma}, D.P., {Welling}, M.: {Auto-Encoding Variational Bayes}. ArXiv
  e-prints  (Dec 2013)

\bibitem{vae_inverseflow}
Kingma, D.P., Salimans, T., Jozefowicz, R., Chen, X., Sutskever, I., Welling,
  M.: Improved variational inference with inverse autoregressive flow. In: Lee,
  D.D., Sugiyama, M., Luxburg, U.V., Guyon, I., Garnett, R. (eds.) Advances in
  Neural Information Processing Systems 29, pp. 4743--4751. Curran Associates,
  Inc. (2016),
  \url{http://papers.nips.cc/paper/6581-improved-variational-inference-with-inverse-autoregressive-flow.pdf}

\bibitem{NADE_autoregressiveDistri_estimator_2011}
Larochelle, H., Murray, I.: The neural autoregressive distribution estimator.
  In: Gordon, G., Dunson, D., Dudík, M. (eds.) Proceedings of the Fourteenth
  International Conference on Artificial Intelligence and Statistics.
  Proceedings of Machine Learning Research, vol.~15, pp. 29--37. PMLR, Fort
  Lauderdale, FL, USA (11--13 Apr 2011),
  \url{http://proceedings.mlr.press/v15/larochelle11a.html}

\bibitem{Rodrigo2018}
Mendoza, R.: {Anomaly Detection with generative models}. Master's thesis,
  {Eindhoven University of Technology} (2018)

\bibitem{LOCI_density_outlier}
Papadimitriou, S., Kitagawa, H., Gibbons, P.B., Faloutsos, C.: Loci: fast
  outlier detection using the local correlation integral. In: Proceedings 19th
  International Conference on Data Engineering (Cat. No.03CH37405). pp.
  315--326 (March 2003). \doi{10.1109/ICDE.2003.1260802}

\bibitem{masked_autoregressiveFlow_2017_artstate}
{Papamakarios}, G., {Pavlakou}, T., {Murray}, I.: {Masked Autoregressive Flow
  for Density Estimation}. ArXiv e-prints  (May 2017)

\bibitem{Radford_DCGAN}
Radford, A., Metz, L., Chintala, S.: Unsupervised representation learning with
  deep convolutional generative adversarial networks. CoRR
  \textbf{abs/1511.06434} (2015), \url{http://arxiv.org/abs/1511.06434}

\bibitem{CADe_medicalImages}
Roth, H.R., Lu, L., Liu, J., Yao, J., Seff, A., Cherry, K.M., Kim, L., Summers,
  R.M.: Improving computer-aided detection using convolutional neural networks
  and random view aggregation. CoRR  \textbf{abs/1505.03046} (2015),
  \url{http://arxiv.org/abs/1505.03046}

\bibitem{AnoGAN}
Schlegl, T., Seeb{\"{o}}ck, P., Waldstein, S.M., Schmidt{-}Erfurth, U., Langs,
  G.: Unsupervised anomaly detection with generative adversarial networks to
  guide marker discovery. CoRR  \textbf{abs/1703.05921} (2017),
  \url{http://arxiv.org/abs/1703.05921}

\bibitem{Ultrasound_smallData}
Shi, J., Zhou, S., Liu, X., Zhang, Q., Lu, M., Wang, T.: Stacked deep
  polynomial network based representation learning for tumor classification
  with small ultrasound image dataset. Neurocomputing  \textbf{194},  87 -- 94
  (2016). \doi{https://doi.org/10.1016/j.neucom.2016.01.074},
  \url{http://www.sciencedirect.com/science/article/pii/S0925231216002344}

\bibitem{MLEToLoss}
Sunehag, P., Trumpf, J., Vishwanathan, S.V.N., Schraudolph, N.N.: Variable
  metric stochastic approximation theory. In: {van Dyk}, D., Welling, M. (eds.)
  Proc.\ 12-th Intl.\ Conf.\ Artificial Intelligence and Statistics (AIstats).
  Workshop and Conference Proceedings, vol.~5, pp. 560--566. Clearwater Beach,
  Florida (2009)

\bibitem{RNADE_autoregressiveDistri_estimator_2013}
{Uria}, B., {Murray}, I., {Larochelle}, H.: {RNADE: The real-valued neural
  autoregressive density-estimator}. ArXiv e-prints  (Jun 2013)

\bibitem{Raymond_2GAN}
Yeh, R., Chen, C., Lim, T.Y., Hasegawa-Johnson, M., Do, M.N.: {Semantic Image
  Inpainting with Deep Generative Models}. arXiv:1607.07539  (2016),
  \url{https://arxiv.org/abs/1607.07539}

\end{thebibliography}
\end{document}